\documentclass[runningheads]{llncs}
\usepackage{esvect}
\usepackage[T1]{fontenc}
\usepackage{graphicx,verbatim}
\usepackage{amsmath}
\usepackage{booktabs}
\usepackage{multirow}
\begin{document}
\title{A Guideline-Aware AI Agent for Zero-Shot Target Volume Auto-Delineation}
%

\makeatletter
\newcommand{\printfnsymbol}[1]{%
\textsuperscript{\@fnsymbol{#1}}%
}
\makeatother

\author{
Yoon Jo Kim\inst{1}\thanks{Equal contribution} \and
Wonyoung Cho\inst{1}\printfnsymbol{1} \and
Jongmin Lee\inst{1} \and
Han Joo Chae\inst{1} \and
Hyunki Park\inst{2} \and
Sang Hoon Seo\inst{2} \and
Jae Myung Noh\inst{2} \and
Kyungmi Yang\inst{2} \and
Dongryul Oh\inst{2}\thanks{Co-corresponding authors} \and
Jin Sung Kim\inst{1}\printfnsymbol{2}
}

\authorrunning{Y.J. Kim et al.}

\institute{Oncosoft Inc., Seoul, South Korea \and
Department of Radiation Oncology, Samsung Medical Center, Sungkyunkwan University School of Medicine, Seoul, South Korea\\
\email{dongryul.oh@samsung.com, jin@oncosoft.io}}

  
\maketitle              
\begin{abstract}
Delineating the clinical target volume (CTV) in radiotherapy involves complex margins constrained by tumor location and anatomical barriers.
While deep learning models automate this process, their rigid reliance on expert-annotated data requires costly retraining whenever clinical guidelines update.
To overcome this limitation, we introduce OncoAgent, a novel guideline-aware AI agent framework that seamlessly converts textual clinical guidelines into three-dimensional target contours in a training-free manner.
Evaluated on esophageal cancer cases, the agent achieves a zero-shot Dice similarity coefficient of 0.842 for the CTV and 0.880 for the planning target volume, demonstrating performance highly comparable to a fully supervised nnU-Net baseline.
Notably, in a blinded clinical evaluation, physicians strongly preferred OncoAgent over the supervised baseline, rating it higher in guideline compliance, modification effort, and clinical acceptability.
Furthermore, the framework generalizes zero-shot to alternative esophageal guidelines and other anatomical sites (e.g., prostate) without any retraining.
Beyond mere volumetric overlap, our agent-based paradigm offers near-instantaneous adaptability to alternative guidelines, providing a scalable and transparent pathway toward interpretability in radiotherapy treatment planning.


\end{abstract}

\section{Introduction}
Radiation therapy requires precise delineation of the clinical target volume (CTV) to encompass microscopic disease while strictly sparing organs at risk (OARs).
Such delineation frequently involves complex, non-isotropic margins determined by tumor location and surrounding anatomical barriers.
As a result, this manual process remains labor-intensive and highly prone to substantial inter-observer variability~\cite{njeh2008,chang2021}.
While deep learning (DL) models like nnU-Net~\cite{isensee2021} have automated some aspects of medical segmentation, they remain ``black boxes'' that require extensive, expert-annotated datasets.
Moreover, these models are inherently rigid: any update to clinical guidelines requires the costly re-annotation of entire datasets and subsequent model retraining.

Recent advances in large language models (LLMs) have given rise to \emph{AI agents}: systems that combine an LLM's reasoning capabilities with external tools to perform complex, multi-step tasks~\cite{wang2024,schick2023}.
An AI agent can interpret a clinical guideline written in natural language, decompose it into an executable plan, and invoke specialized tools---such as OAR segmentation models and morphological operators---to autonomously construct patient-specific target volumes.

This paradigm offers three key advantages over data-driven DL: (1)~\emph{no requirement for expert annotations}, since the delineation logic is directly derived from the guideline text rather than learned from examples; (2)~\emph{instant adaptability} to evolving guidelines or institutional variations, as merely updating the textual input is sufficient to reflect new clinical protocols; and (3)~\emph{interpretability}, as the agent produces a human-readable plan before execution that clinicians can review and approve.

In this work, we present OncoAgent, an AI agent framework for universal, guideline-aware target volume auto-delineation and evaluate it on esophageal cancer patients.
Our contributions are threefold:

\begin{enumerate}
    \item We propose a general, two-phase agentic architecture---\emph{planning} and \emph{execution}---that converts free-text clinical guidelines into three-dimensional target volumes using pre-trained OAR segmentation models and geometric operation tools in a completely training-free manner.
    \item Using a mid-thoracic esophageal cancer dataset, we demonstrate that OncoAgent achieves performance highly comparable to a fully supervised nnU-Net baseline, while inherently ensuring superior anatomical safety and strict guideline compliance.
    \item We validate the framework's broad applicability through cross-guideline transferability experiments, proving that OncoAgent adapts instantly to alternative esophageal protocols and generalizes zero-shot to other anatomical sites (e.g., prostate).
\end{enumerate}

\section{Method}

\subsection{Overview}

\begin{figure}[t]
  \centering
  \includegraphics[width=\textwidth]{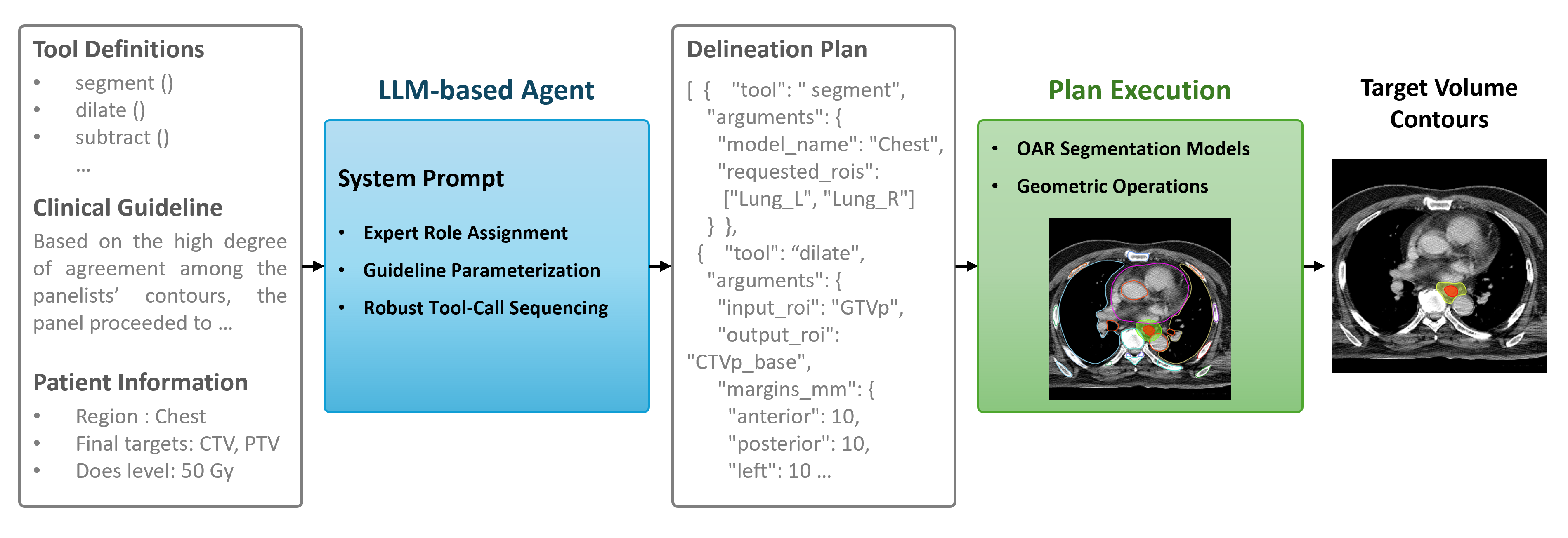}
  \caption{Overview of the proposed OncoAgent framework for zero-shot target volume auto-delineation.}
  \label{fig:overview}
\end{figure}

OncoAgent automatically converts textual clinical guidelines into 3D target volumes in a zero-shot manner (Fig.~\ref{fig:overview}).
Given tool definitions, a clinical guideline, and patient information, a large language model (LLM) interprets the raw text guideline to formulate a delineation plan---an ordered sequence of tool calls expressed as a JSON array.
The plan is validated against the tool-call schema~\cite{geng2025}; if violations are detected, the LLM iteratively self-refines~\cite{madaan2023selfrefine} until conformance.
A deterministic execution engine then dispatches each call on the patient's CT and gross tumor volume (GTV) contours, invoking pre-trained OAR segmentation models and geometric operations to produce the final target volume.

\subsection{System Prompt Strategies}
\label{sec:prompt}

The system prompt defines the core logic of the agent, structuring raw text guidelines into deterministic, actionable tool-call sequences.
It encodes three complementary prompting strategies (Fig.~\ref{fig:prompt}) that enable OncoAgent to operate effectively in a zero-shot setting.

\paragraph{Clinician Role Assignment.}
Following the role prompting paradigm~\cite{wang2024}, the system prompt assigns the LLM the persona of an expert radiation oncology agent.
To ensure deterministic, guideline-compliant delineation, the prompt encodes the standard target volume definitions~\cite{icru1999,shusharina2020} as fixed computational templates following the program-aided reasoning paradigm~\cite{gao2023pal}:
\begin{equation}
  \text{CTV} = (\text{GTV} \oplus \mathbf{m}_{\text{ctv}}) \setminus \text{OARs},\qquad
  \text{PTV} = \text{CTV} \oplus \mathbf{m}_{\text{ptv}},
  \label{eq:tv}
\end{equation}
where $\mathbf{m}$ denotes a direction-specific margin vector.
These domain-specific templates guide the agent to produce delineation plans that are explicitly aligned with clinical guidelines.

\paragraph{Guideline Parameterization.}
Clinical guidelines often exhibit terminological variation and present margin values as ranges.
The system prompt provides example OAR aliases and a list of segmentable structures, enabling the LLM to map guideline terms to model-specific OAR names (e.g., ``Lung''~$\to$~\texttt{Lung\_L}, \texttt{Lung\_R}).
When guidelines specify ranges (e.g., ``5--10\,mm''), OncoAgent resolves them using patient-specific context (e.g., dose level, physician preferences).

\paragraph{Robust Tool-Call Sequencing.}
Adapted from algorithmic prompting~\cite{zhou2023}, the system prompt specifies a step-by-step execution order of segmentation and geometric tools, such as \texttt{segment}, \texttt{dilate}, and \texttt{subtract}, to avoid delineation errors.
Additionally, consistent naming conventions ensure correct ROI referencing throughout the generated plan.
If the generated plan is invalid, the LLM iteratively self-refines the plan until it works.

\begin{figure}[t]
  \centering
  \footnotesize
  \setlength{\tabcolsep}{4pt}
  \renewcommand{\arraystretch}{1.15}
  \begin{tabular}{@{}p{0.95\textwidth}@{}}
    \toprule
    \textbf{System Prompt} (simplified) \\
    \midrule

    \textbf{Clinician Role Assignment} \\
    \quad $\bullet$\; ``You are an expert radiation oncology guideline-to-implementation agent.'' \\
    \quad $\bullet$\; Mandatory template:
      $\text{CTV} = (\text{GTV} \oplus \mathbf{m}_{\text{ctv}}) \setminus \text{OARs}$;\;
      $\text{PTV} = \text{CTV} \oplus \mathbf{m}_{\text{ptv}}$ \\[4pt]

    \textbf{Guideline Parameterization} \\
    \quad $\bullet$\; \emph{OAR alias examples}: \\
    \quad\quad Lung $\to$ \texttt{Lung\_L}, \texttt{Lung\_R};\quad
           Kidney $\to$ \texttt{Kidney\_L}, \texttt{Kidney\_R};\quad
           VBs $\to$ \texttt{VB\_whole} \\
    \quad $\bullet$\; Multi-OAR aliases: apply operations per component, then union \\
    \quad $\bullet$\; Resolve margin ranges using patient context (e.g., dose level, physician preferences), producing patient-specific delineation plans \\[4pt]

    \textbf{Robust Tool-Call Sequencing} \\
    \quad 1. Query existing ROIs and segmentation models \\
    \quad 2. \texttt{segment}: delineate required OARs \\
    \quad 3. \texttt{union}: OARs $\to$ unified exclusion mask \\
    \quad 4. \texttt{dilate}: GTV $\to$ CTV base \\
    \quad 5. \texttt{subtract}: CTV base $-$ exclusion mask $\to$ final CTV \\
    \quad 6. \texttt{dilate}: CTV $\to$ PTV \\
    \quad 7. Validate against schema; self-refine if violations found \\

    \bottomrule
  \end{tabular}
  \caption{Simplified system prompt. The three strategies are detailed in Sec.~\ref{sec:prompt}.}
  \label{fig:prompt}
\end{figure}

\subsection{Plan Execution}
\label{sec:execution}

The execution engine receives the validated tool-call sequence and dispatches each call sequentially, passing the output of each step as input to subsequent ones.
OncoAgent has access to two classes of tools (Fig.~\ref{fig:prompt}):
a \textit{segmentation tool} that invokes pre-trained deep learning models to delineate anatomical structures on the patient CT, and \textit{geometric operation tools} that apply direction-specific morphological and Boolean operations.
After all planned calls have been executed, post-processing---contour smoothing and hole-filling---is applied to produce the final target volume.
\section{Results}
\subsection{Experimental Setup}

\paragraph{Dataset.}
We retrospectively collected planning CT scans from 40 patients with mid-thoracic esophageal cancer, provided by Samsung Medical Center under the approval of the Data Review Board (DRB No. D2025-0006) and the Institutional Review Board (IRB No. SMC 2024-11-122).
For each case, an experienced radiation oncologist delineated ground truth (GT) contours for the GTV, CTV, and PTV in accordance with the IJROBP expert consensus guideline~\cite{wu2015}.
All CT images were anonymized and exported from the Eclipse treatment planning system.

\paragraph{Comparative Methods.}
We compared OncoAgent against three supervised DL baselines: 
(1) nnU-Net~\cite{isensee2021}, a robust, self-configuring medical image segmentation framework; 
(2) DDAU-Net~\cite{yousefi2021}, a state-of-the-art dual-attention model for esophageal target volume segmentation; 
and (3) nnU-Net (GTV Prior), a strong baseline utilizing both CT images and GT GTV masks as a 2-channel input for GTV-guided segmentation~\cite{kihara2023}. The DL models were trained on 32 cases, and both the DL models and OncoAgent were evaluated on the remaining 8 test cases.

\paragraph{Evaluation Metrics and Clinical Assessment.}
Quantitative performance was evaluated using Dice Similarity Coefficient (DSC), Mean Surface Distance (MSD), Sensitivity, and Precision, with statistical significance determined by a paired t-test ($p < 0.05$).
To assess planning accuracy, we additionally used \textit{Tool Call F1}, the set-level F1 score between the agent-generated and a manually authored reference tool-call sequence.

For clinical assessment, two senior radiation oncologists (17 and 20 years' experience) blindly rated the 8 test cases generated by nnU-Net (GTV Prior) and OncoAgent. 
Using a 5-point Likert scale, they evaluated Guideline Compliance, Modification Effort, and Clinical Acceptability, respectively.
Qualitative analysis assessed anatomical violations via visual comparison.

\paragraph{Implementation Details.}
The system integrates GPT-5.2~\cite{openai2025gpt5} as its reasoning engine, utilizing a pre-trained deep learning model for OAR segmentation and the VTK library~\cite{schroeder2006} for geometric operations.
This end-to-end workflow achieved an average execution time of 1.74 minutes per patient, requiring a mean of 1.13 LLM inference calls per case to generate a complete set of target contours.

\subsection{Quantitative Evaluation}
The quantitative performance of OncoAgent is presented in Table~\ref{tab:detailed_metrics}.
While baselines relying solely on CT images (DDAU-Net, nnU-Net) struggled with invisible, guideline-derived CTV boundaries, incorporating GTV prior knowledge---either via supervised multi-channel inputs (nnU-Net with GTV Prior) or zero-shot guideline reasoning (OncoAgent)---yielded substantial performance gains.

Notably, OncoAgent demonstrated performance highly comparable to the fully supervised state-of-the-art model, nnU-Net (GTV Prior), despite achieving this entirely zero-shot without the annotated training cases required by the DL baselines.
Both models maintained high overall performance (DSC $\ge$ 0.84) and sub-voxel distance errors (MSD $\approx$ 1.0\,mm).
As indicated by the bold values in Table~\ref{tab:detailed_metrics}, there were no statistically significant differences ($p \ge 0.05$) between the two models across primary metrics like DSC and MSD.
While the supervised model showed slightly higher Precision, OncoAgent achieved comparable or superior Sensitivity, yielding similar overall performance.

\begin{table}[!t]
\centering
\caption{Quantitative evaluation of target volume segmentation (mean $\pm$ SD). MSD is in mm. Bold: best or no significant difference ($p \geq 0.05$) from best.}
\label{tab:detailed_metrics}
\footnotesize
\setlength{\tabcolsep}{1pt}
\renewcommand{\arraystretch}{1.1}
\begin{tabular}{@{}l l c c c c@{}}
\toprule
\textbf{Class} & \textbf{Method} & \textbf{DSC $\uparrow$} & \textbf{MSD $\downarrow$} & \textbf{Sens. $\uparrow$} & \textbf{Prec. $\uparrow$} \\
\midrule
\multirow{4}{*}{CTV}
& DDAU-Net & 0.463${\pm}$0.107 & 8.81${\pm}$3.88 & 0.345${\pm}$0.103 & 0.762${\pm}$0.143 \\
& nnU-Net & 0.627${\pm}$0.130 & 7.82${\pm}$6.90 & 0.683${\pm}$0.071 & 0.594${\pm}$0.183 \\
& nnU-Net (GTV Prior) & \textbf{0.862${\pm}$0.026} & \textbf{0.95${\pm}$0.22} & \textbf{0.845${\pm}$0.053} & \textbf{0.882${\pm}$0.028} \\
& OncoAgent (Ours) & \textbf{0.842${\pm}$0.026} & \textbf{1.06${\pm}$0.15} & \textbf{0.884${\pm}$0.029} & 0.807${\pm}$0.053 \\
\midrule
\multirow{4}{*}{PTV}
& DDAU-Net & 0.517${\pm}$0.088 & 7.95${\pm}$3.46 & 0.398${\pm}$0.098 & 0.785${\pm}$0.119 \\
& nnU-Net & 0.659${\pm}$0.118 & 8.77${\pm}$7.16 & 0.723${\pm}$0.059 & 0.620${\pm}$0.171 \\
& nnU-Net (GTV Prior) & \textbf{0.893${\pm}$0.021} & \textbf{1.02${\pm}$0.25} & 0.873${\pm}$0.045 & \textbf{0.916${\pm}$0.020} \\
& OncoAgent (Ours) & \textbf{0.880${\pm}$0.015} & \textbf{1.12${\pm}$0.14} & \textbf{0.922${\pm}$0.027} & 0.843${\pm}$0.033 \\
\bottomrule
\end{tabular}
\end{table}

\subsection{Clinical Assessment and Qualitative Analysis}

The blinded physician survey is summarized in Table~\ref{tab:survey}, showing a strong clinical preference for OncoAgent.
OncoAgent outperformed the nnU-Net (GTV Prior) model across all criteria, especially in Guideline Compliance ($4.06$ vs. $3.56$) and Modification Effort ($4.12$ vs. $3.44$).

\begin{figure}[t]
\centering
\includegraphics[width=\textwidth]{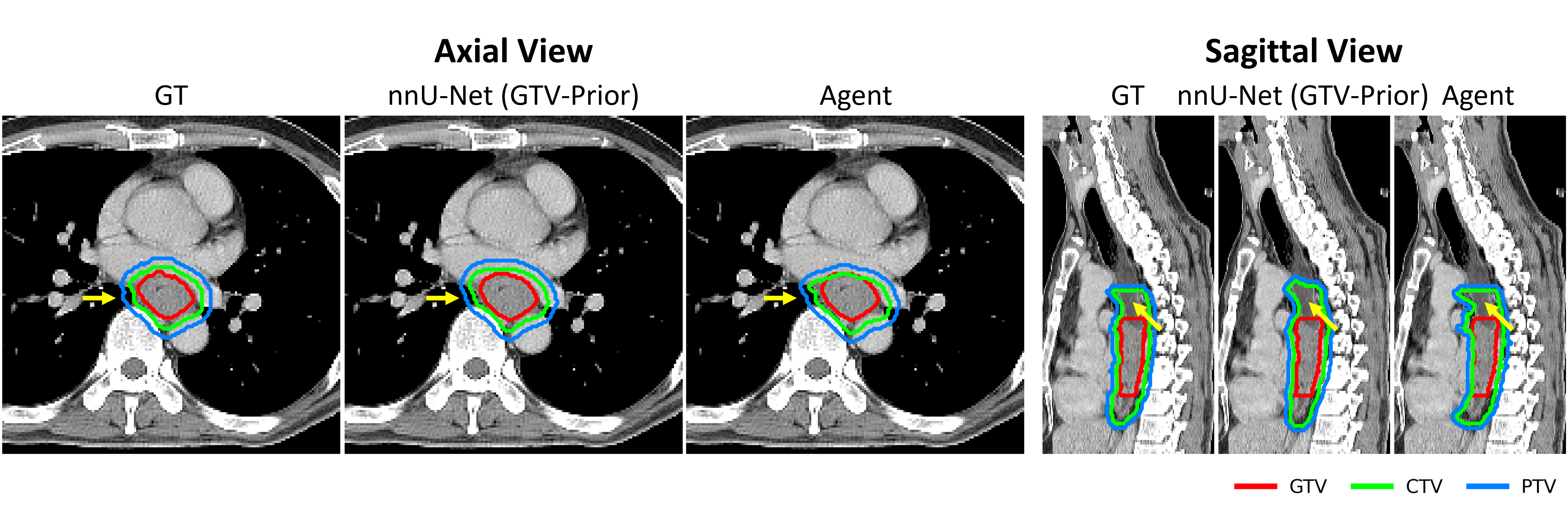}
\caption{Qualitative comparison of nnU-Net (GTV Prior) and OncoAgent in axial and sagittal views. Yellow arrows highlight areas of interest.}
\label{fig:qualitative}
\end{figure}

Visual comparisons between the manual GT, nnU-Net (GTV Prior), and OncoAgent results are illustrated in Fig.~\ref{fig:qualitative}.
In the axial view (yellow arrows), the DL model often invades adjacent OARs (e.g., lung) as it lacks explicit end-points.
On the other hand, OncoAgent strictly applies explicit boolean exclusion rules thereby avoiding anatomical violations; although this rigid subtraction mechanism led to a ``slightly over-eroded'' appearance near the heart compared to the GT, it successfully avoided critical OAR overlap and resulted in a safer target volume.
Furthermore, in the sagittal view, OncoAgent accurately matched the GT by strictly adhering to the specified longitudinal ($z$-axis) margins, while the DL model arbitrarily over-expanded by approximately $6$\,mm superiorly.

\begin{table}[t]
\centering
\caption{Blinded physician survey (5-point Likert scale). \textbf{Bold}: highest.}
\label{tab:survey}
\small
\setlength{\tabcolsep}{4pt}
\renewcommand{\arraystretch}{1.1}
\begin{tabular}{@{}llcc@{}}
\toprule
\textbf{Evaluation Criteria} & \textbf{Method} & \textbf{Mean $\pm$ SD} & \textbf{Ratings $\geq 4$ (\%)} \\
\midrule
\multirow{2}{*}{Guideline Compliance}
& nnU-Net (GTV Prior) & 3.56 $\pm$ 0.51 & 56.2\% \\
& \textbf{OncoAgent (Ours)} & \textbf{4.06 $\pm$ 0.68} & \textbf{81.2\%} \\
\midrule
\multirow{2}{*}{Modification Effort}
& nnU-Net (GTV Prior) & 3.44 $\pm$ 0.63 & 37.5\% \\
& \textbf{OncoAgent (Ours)} & \textbf{4.12 $\pm$ 0.89} & \textbf{68.8\%} \\
\midrule
\multirow{2}{*}{Clinical Acceptability}
& nnU-Net (GTV Prior) & 3.38 $\pm$ 0.72 & 50.0\% \\
& \textbf{OncoAgent (Ours)} & \textbf{3.81 $\pm$ 0.75} & \textbf{75.0\%} \\
\bottomrule
\end{tabular}
\end{table}

\subsection{Zero-Shot Generalization and Cross-Guideline Adaptability}
Unlike DL models requiring retraining for guideline updates, OncoAgent adapts immediately.
We evaluated it on alternative esophageal guidelines (CROSS Trial~\cite{cross_protocol}, JASTRO 2024~\cite{jastro2024}) and extended it to prostate cancer (RTOG 0126~\cite{rtog0126}) without retraining.
As shown in Table~\ref{tab:cross_guideline}, OncoAgent attained high Tool Call F1 scores (0.73--1.00) across esophageal guidelines, with a lower score (0.64) for prostate due to unfamiliar anatomical terms (e.g., ``SV\_prox1cm''), confirming cross-guideline and cross-site generalizability without task-specific fine-tuning.

\begin{table}[t]
\centering
\caption{Zero-shot transferability across different clinical guidelines and a novel site (prostate).}
\label{tab:cross_guideline}
\begin{tabular}{@{}lc@{}}
\toprule
\textbf{Target Site \& Guideline} & \textbf{Tool Call F1} $\uparrow$ \\
\midrule
Esophagus (IJROBP Expert Consensus$^\dagger$~\cite{wu2015}) & 1.00 \\
Esophagus (JASTRO 2024 Guideline~\cite{jastro2024}) & 0.78 \\
Esophagus (CROSS Trial~\cite{cross_protocol}) & 0.73 \\
Prostate (RTOG 0126~\cite{rtog0126}) & 0.64 \\
\bottomrule
\end{tabular}\\[2pt]
{\footnotesize $^\dagger$Primary guideline used in the main evaluation.}
\end{table}

\section{Discussion and Conclusion}
\subsection{Discussion}
We have presented OncoAgent, the first AI agent framework for guideline-aware target volume auto-delineation, evaluated on mid-thoracic esophageal cancer.
This approach shifts the focus from \emph{learning CTV from annotated data} to \emph{reasoning about CTV from clinical guidelines}, yielding three distinct advantages.

\paragraph{Training-Free Deployment.}
OncoAgent requires zero CTV-labeled training data, relying instead on a clinical guideline document and pre-trained OAR segmentation models.
Despite this zero-shot approach, it achieved performance comparable to the fully supervised nnU-Net (GTV Prior), with no statistically significant differences in primary metrics such as DSC and MSD.
Furthermore, in blinded clinical evaluations, physicians demonstrated a strong preference for OncoAgent over the supervised baseline, rating it higher in both Guideline Compliance (4.06 vs. 3.56) and Modification Effort (4.12 vs. 3.44).

\paragraph{Instant Guideline Adaptability.}
When guidelines evolve or institutional protocols vary, OncoAgent adapts through simple document updates, bypassing the need for costly re-annotation or model retraining.
We demonstrated this by transitioning between the IJROBP expert consensus guideline~\cite{wu2015}, CROSS~\cite{cross_protocol}, and the JASTRO 2024 guideline~\cite{jastro2024}, 
producing appropriately distinct CTV contours without any additional training.
This capability directly addresses a critical challenge in clinical practice, where conventional DL models often lose their clinical applicability with each guideline update.

\paragraph{Interpretability and Auditability.}
OncoAgent's intermediate plan---specifying margins, excluded OARs, and execution steps in human-readable form---enables clinicians to review and approve the delineation logic before contour generation.
This transparency is critical for regulatory compliance and clinical trust, and is fundamentally absent in end-to-end DL approaches.

\paragraph{Limitations.}
First, OncoAgent's performance depends on the quality of the underlying OAR segmentation model; however, recent advancements in robust auto-segmentation substantially mitigate this concern.
Second, as with any LLM-based system, OncoAgent is susceptible to semantic misinterpretation and hallucination\allowbreak{}---for instance, misapplying a guideline clause to an inappropriate anatomical context---\allowbreak{}and cannot capture implicit clinical knowledge absent from the guideline text.
Iterative self-refinement rejects structurally invalid plans before execution, as empirically supported by high Tool Call F1 scores (0.73--1.00) across diverse guidelines; nevertheless, clinician review of the interpretable execution plan remains essential before clinical use.
Third, while OncoAgent's strict guideline compliance yields geometrically precise contours, this may lack the pragmatic safety margins that physicians manually draw to accommodate esophageal motion from respiration, swallowing, and cardiac pulsation; future work should integrate adaptive margin strategies to reconcile geometric precision with motion robustness.
Fourth, this study is limited to 40 patients from a single institution; multi-center validation, extension to other tumor sites (e.g., head and neck, cervical cancer), and clinician feedback loops are important future directions.

\subsection{Conclusion}
OncoAgent demonstrates that LLM-based reasoning, combined with pre-trained segmentation tools and geometric operation tools, 
can achieve clinically competitive target volume delineation without any expert-annotated training data.
Notably, in a blinded physician survey, OncoAgent demonstrated strong clinical preference over the supervised baseline, achieving higher ratings in Guideline Compliance, Modification Effort, and Clinical Acceptability.
This framework bridges the gap between text-based clinical guidelines and automated target delineation, offering a transparent and adaptable alternative to conventional data-driven deep learning.

    

\begin{credits}
\subsubsection{\ackname}
The authors would like to thank Samsung Medical Center for providing the clinical data used for validation in this study. This study was approved by the Institutional Review Board of Samsung Medical Center (DRB No. D2025-0006, IRB No. SMC 2024-11-122).

\subsubsection{\discintname}
Y.J.\ Kim, W.\ Cho, J.\ Lee, and H.J.\ Chae are employees of Oncosoft Inc. J.S.\ Kim is a CEO of Oncosoft Inc. The remaining authors have no competing interests to declare.
\end{credits}

\clearpage

%
%
%
\bibliographystyle{splncs04}
\bibliography{references}

\end{document}